\documentclass[10pt,journal]{IEEEtran}
\usepackage{amsmath,amsfonts}
\usepackage{algorithmic}
\usepackage{array}
\usepackage{textcomp}
\usepackage{stfloats}
\usepackage{url}
\usepackage{verbatim}
\usepackage{graphicx}
\usepackage{subfigure}
\usepackage{multirow}
\DeclareMathOperator*{\argmax}{arg\,max}
\def\BibTeX{{\rm B\kern-.05em{\sc i\kern-.025em b}\kern-.08em
    T\kern-.1667em\lower.7ex\hbox{E}\kern-.125emX}}
\usepackage{balance}
\begin{document}
\title{A multi-purpose automatic editing system based on lecture semantics for remote education}
\author{Panwen Hu, Rui Huang
\thanks{Panwen Hu and Rui Huang are with The Chinese University of Hong Kong, Shenzhen, China (e-mail: panwenhu@link.cuhk.edu.cn; yongzhang@link.cuhk.edu.cn; ruihuang@cuhk.edu.cn).}}


\maketitle

\begin{abstract}
Remote teaching has become popular recently due to its convenience and safety, especially under extreme circumstances like a pandemic.
However, online students usually have a poor experience since the information acquired from the views provided by the broadcast
platforms is limited. One potential solution is to show more camera views simultaneously, but it is technically challenging and distracting for the
viewers. Therefore, an automatic multi-camera directing/editing system, which aims at selecting the most concerned view at each time instance
to guide the attention of online students, is in urgent demand. However, existing systems mostly make simple assumptions and
focus on tracking the position of the speaker instead of the real lecture semantics, and therefore have limited capacities to deliver
optimal information flow. To this end, this paper proposes an automatic multi-purpose editing system based on the lecture semantics, which can both direct the multiple video streams for real-time broadcasting and edit the optimal video offline for review purposes.
Our system directs the views by semantically analyzing the class events while following the professional directing rules,
mimicking a human director to capture the regions of interest from the viewpoint of the onsite students. We conduct both qualitative
and quantitative analyses to verify the effectiveness of the proposed system and its components.
\end{abstract}


\section{Introduction}
\begin{figure}[htbp]
  \flushleft
  \includegraphics[scale=0.45]{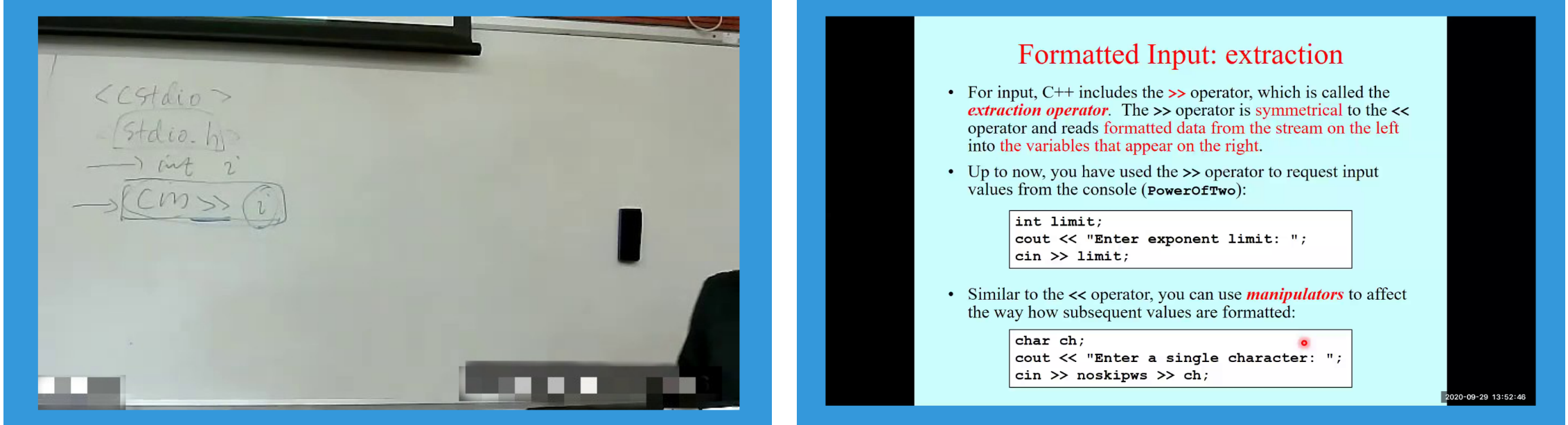}
  \caption{The remote students can watch and manually switch between only two close-up views when taking classes online through online teaching software, e.g., Zoom.}
  \label{fig:zoomview}
  \vspace{-12pt}
\end{figure}

Mixed-mode or hybrid teaching with both onsite and online students has become a popular teaching practice during the pandemic situation, and it also provides a way to spread knowledge and promote education fairness around the world. Nowadays, students who cannot attend the onsite lectures for various reasons can still participate through video conferencing platforms online or watch the recordings offline. Nonetheless, the information and experiences received by these students are inferior to those received by the onsite students. One reason is that the information conveyed through the views provided by the platform is usually very limited, as shown in Fig.\ref{fig:zoomview}. The students cannot acquire the entire events from different views freely, and staring at the same view for a long time may cause mental stress\cite{Jahrir2020}. On the other hand, if the platforms provide the students with many different views, it is both technically challenging (issues with synchronization, bandwidth, etc.) and the multiple video sources are difficult to browse through, so the remote students still have to manually select the view of interest during the class.

\begin{figure*}[htbp]
  \centering
  \includegraphics[scale=0.4]{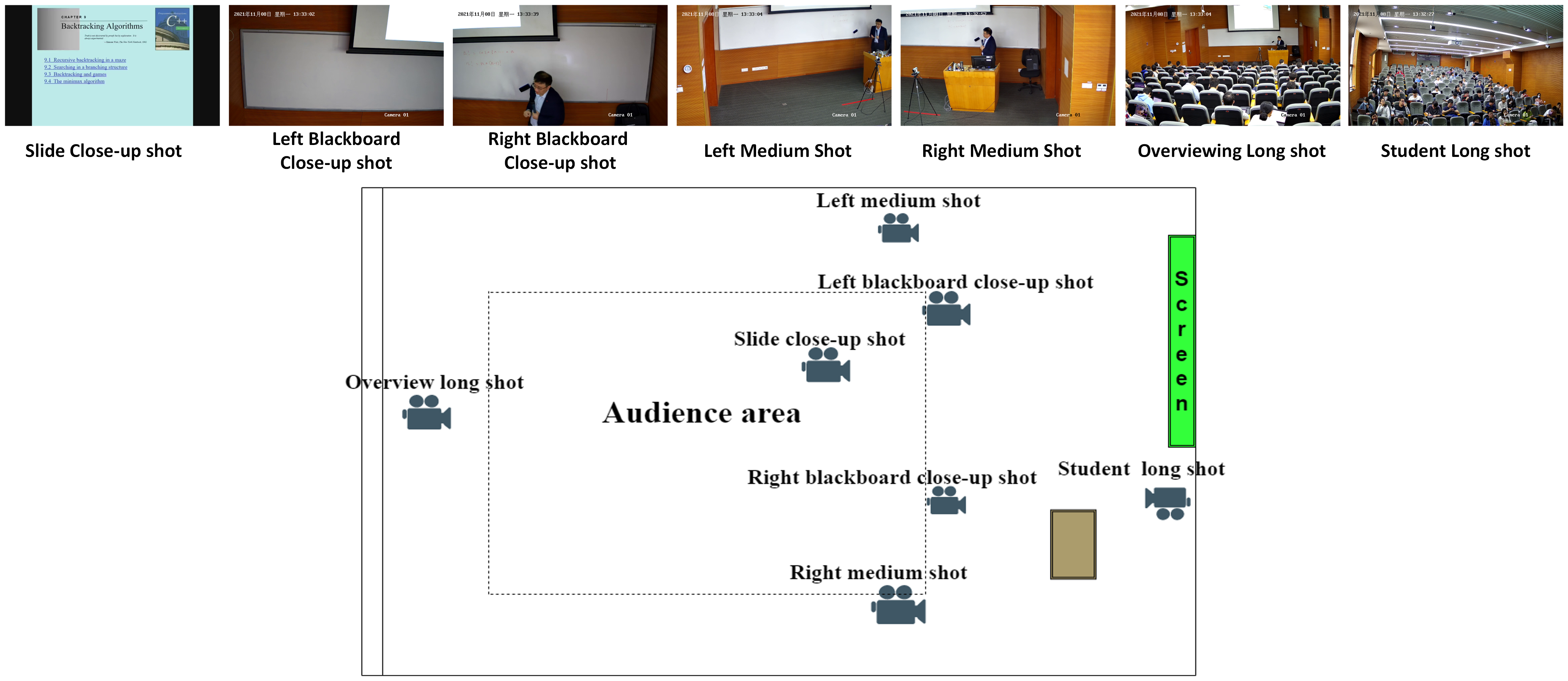}
  \caption{The illustration of our multi-view teaching environment. There are seven video streams, including close-up shots, medium shots, and long shots. Different shots can be used to convey different information..}
  \label{fig:ourview}
\end{figure*}

Recently, a few automatic lectures recording systems \cite{Lampi2007,Zhang2008,Chou2010,Hulens2014} have been proposed. However, these systems focus on automatically adjusting the camera viewpoint to capture the content of interests during the classes, instead of editing multiple video streams together. What the remote student can watch from these systems is either a shot displaying the speaker and his/her surroundings, or a set of raw video streams that require the students to switch manually. It may distract the students and cause information lapses. Therefore, in this paper, we propose an automatic multi-view editing system for lecture videos, which can process more diverse views and automatically edit/switch views based on class semantics. A few similar systems have been proposed by Rui et al. \cite{Liu2001, Rui2004} and Wang et al. \cite{Wang2007}. Their systems adopted the Finite State Machine (FSM) as the editing model where each state corresponds to a camera view, and use the tracked positions and gestures of the speakers as the primary cues to trigger the state transitions. These systems typically contain four cameras and assume that the speakers can directly interact with the projector screen and the blackboard, so the positions of the hand or body are the regions of interest. Whereas, for the large classroom or reporting hall, as shown in Fig.\ref{fig:ourview} which is an extended scene from previous works and used as an experimental scene by our system, these low-level cues are not always effective in representing the focus since students shift their attention according to the events in class instead of where the teacher is. For example, the attention would focus on the slide view instead of the speaker when the speaker flips the slides through a computer. Moreover, the editing rules are hard-coded in the FSM framework employed, which makes the resultant videos too predictable \cite{Aerts2016} and limits the capacity of the system to embed new events and new rules. As a result, it will be difficult for the users to adjust the styles of the generated videos according to their preferences.

In this work, we propose a semantics-based automatic editing system with a computational framework. Unlike previous studies \cite{Bianchi1998, Mukhopadhyay1999, Liu2001, Rui2004} that assume the speaker's positions are the attention regions for the remote students, our system firstly analyzes the semantics of video contents to assess the focus scores of different shots. We observe that the student's attention is dominated by some special events in class. For example, The students would focus on the content of the blackboard when the teacher is writing something, rather than the teacher himself. As the purposes of different shots vary, different semantics analysis methods are proposed to assess the shots based on their functions, e.g., a writing event recognizer is proposed to assess the blackboard close-up shot, and more details are introduced in Sec.\ref{sub:incentives}. In addition, we also take general cinematographic rules \cite{Scott1985, Germeys2007} into consideration to improve the viewing experience. Unlike the previous systems that hard-code editing rules, our system converts the declarative cinematographic rules into computational expressions, as discussed in Sec.\ref{sub:rule expression}.


Besides focus assessment, the editing framework also plays an important role. To improve flexibility and optimality, we propose a multi-purpose optimization-based editing framework. Unlike previous studies that select the shots based on a set of predefined rules, this framework integrates the editing rules, e.g, shot duration, as soft constraints, which allows the system breaks the cycle of rigidity if necessary. For example, if the teacher has been writing something for a relatively long time, the system should stay in the close-up view. In contrast, the rigid shot duration constraint in the rule-based system will motivate a switch to a new view. In addition, our system is multi-purpose and allows users to choose the online mode (live broadcast purpose), offline mode (editing purpose), or a balance between them (look-ahead broadcast purpose) by adjusting a look-ahead duration. More details will be discussed in \ref{sub:selection method}.

To summarize, our contributions mainly include the following aspects:
\begin{enumerate}
  \item Firstly, we propose several practical class semantics analysis methods to assess the attention of shots. To the best of our knowledge, this study is the first attempt to explore video semantics to guide the editing of lecture videos. To evaluate the semantics analysis methods and the proposed editing system, we build a dataset by collecting synchronized multi-view videos from real classes and annotating the writing event. We will make this dataset public to promote the research in this direction.

  \item We further develop a multi-purpose optimization-based editing framework, in which the general editing rules are treated as soft constraints to achieve an optimal solution, and the users can choose different modes by simply adjusting the look-ahead duration.

  \item Qualitative and quantitative analyses have been conducted on the collected dataset to demonstrate the effectiveness of the proposed system and its components. Moreover, in order to compare the real user experience of different systems, we also conduct a user study to assess our system. 
\end{enumerate}

\section{Related Work}
The terminology of video editing is fluctuating in different areas, and in this paper, video editing refers to the process of shot selection from multiple videos along the timeline, instead of changing the contents of video frames (like image editing). In that sense, automatic video editing systems are sometimes called mashup \cite{Shrestha2010} or montage \cite{Bloch1988} systems, which have attracted much attention from the multimedia and computer vision communities. 

This section will briefly review the relevant editing systems. According to the timeline relationship between the raw videos and the resultant videos, we categorize exiting editing systems into two types following previous study \cite{Saini2018}, asynchronous and synchronous systems. The asynchronous systems often require scripts to specify the scene, and the timelines of their resultant videos do not correspond to the time of inputted videos. Video summarization is also a kind of asynchronous editing, but it focuses on extracting the representative parts from a single video instead of multiple video streams, so we exclude video summarization in this section. The synchronous systems, e.g., live broadcasting systems, take as inputs multiple synchronized video streams, and the resultant videos will cover the whole timelines of inputted videos. Our system is a synchronous system whose resultant video has a consistent timeline with the lecture.

\subsection{Asynchronous editing systems}
The montage system by Bloch et al.\cite{Bloch1988} firstly implements automatic film editing. It takes the annotated video rushes as inputs and generates the film sequences for the specified scenario. The constraints on gaze, motion, and positions of the actors borrowed from film theory are applied during production. IDIC \cite{Sack1994} follows Bloch's system with another attempt to generate film from annotated movie shots automatically. IDIC formulates montage as a planning problem and defines a set of operators based on film theory to select and plan video shots for the given story. Christianson et al. \cite{Christianson1996} introduce the Declarative Camera Control Language (DCCL) for generating idiom-based film sequences. Specifically, DCCL uses a particular set of film idioms for editing a particular scene. For example, it uses the conversation idioms for filming the conversation scene, the fighting idioms for filming the fighting scene, etc. Finally, a hierarchical film tree consisting of the idioms for each scene is built to select the shot for the given scene. Unlike previous work, Darshak \cite{Jhala2010} took the extra causal links and ordering constraints as input, besides the story plan and annotated videos. A hierarchical partial-order planner is responsible for selecting the shot sequences that satisfy the constraint and achieve the inputted story goals. Instead of selecting video shots based on the idioms and constraints, Some systems \cite{Elson2007,Galvane2015} formulate the selections of shots as an optimization problem. It first segments an input script into a sequence of scenes. Aesthetic constraints such as location constraints, blocking constraints, etc., are proposed to compute the quality score of shots for each scene. Finally, the dynamic programming method determines the shot sequence that achieves the highest score. Although these systems are successful attempts at editing animated videos, their success heavily relies on the annotations of video content and camera parameters in the virtual world.

Recently, editing real-world videos has also been studied. Leake et al. \cite{Leake2017} propose a computational video editing framework for dialogue scenes. The video annotations required by the film-editing idioms, e.g., the face position of the actors, and the speaker visibility, are generated using advanced computer vision techniques. Finally, a Hidden Markov Model (HMM) and the Viterbi algorithm are employed to compose the film for the script. Moreover, Wang et al. \cite{Wang2019} propose a method for generating a video montage illustrating a given narration. For each sentence in the text, their system retrieves the best-matched shot from the video based using the visual-semantic matching technique.

On the one hand, due to the domain gap, the methods to collect the visual elements for editing are not applicable to the lecture scene. Technically, these systems are not fully automated when analyzing video semantics but require manual annotations. On the other hand, these asynchronous systems always edit videos based on a given script, which specifies the content and the temporal relationships of shots. As a result, the edited videos do not always hold a complete timeline of the raw input videos. However, for the scenes like lectures broadcasting, and sport match broadcasting, the scripts are not available due to the immediacy and the high dynamics, and the timeline of the resultant video should just cover the whole activity, i.e., the lecture, without redundancy or deficiency. To this end, our focus is on editing multiple synchronous lecture video streams together using class semantics, and the generated video has a consistent timeline with the input videos.

\subsection{Synchronous editing systems}
Synchronized editing also has drawn much attention due to its wide applications. For example, previous studies \cite{Ariki2006,Wang2008,Kaiser2012,Li2020, pan2021} have proposed systems for live broadcasting soccer game. In this system, the motion of players, the location of the ball \cite{Wang2008,Li2020}, people detection, and saliency model \cite{Niamut2013} are used as intermediate representations for high-level event detection, which is used to evaluate the importance of each camera view. The system by Quiroga et al.\cite{Quiroga2020} is developed for automatically broadcasting basketball games, where the locations of the ball and players, and the mapping relationship between the frame and the court are jointly used to recognize the game state. Besides broadcasting of sports events, the synchronized editing for concert recording \cite{Shrestha2010}, performance video \cite{Saini2012, Wu2015,Bano2016}, social video \cite{Arev2014}, and surveillance video \cite{Hu2021}, etc. have been explored as well. Compared to these types of videos, the lecture videos that our system explores lie in a significantly different content domain, and the editing rules used are not compatible. Therefore, directly converting these systems to accommodate the lecture scene is non-trivial, even though changing the content measurements.  

A few pieces of literature \cite{Bianchi1998, Mukhopadhyay1999, Liu2001, Rui2004, Wang2007} have attempted to edit lecture videos. The tracked body positions \cite{Rui2004, Mavlankar2010, Winkler2012}, the gesture \cite{Wang2007} or the head positions \cite{Chou2010} are always considered the most important cues to switch or plan the cameras. Occasionally, additional features such as gaze direction \cite{Hulens2014}, and the position relationship between the lecturer and the chalkboard \cite{Pang2010} are incorporated. However, we observe that these position-aware representations are insufficient to determine the student's attention in class. Some important events, e.g., slide flips with a computer, have little relation to the positions of the speaker. To this end, our system comes up with a few practical video semantics analysis methods, as well as the computational expressions of empirical editing rules, to guide the editing. 

On the other hand, most existing editing frameworks imitate the human director by applying the predefined selection rules \cite{Zhang2008, Chen2014}, or the script language \cite{Kaiser2012}. For example, Machnicki et al.\cite{Machnicki2001} describe that after showing the close-up speaker for 1 minute, the system should switch to the stage view and show that for 15 seconds. Some other frameworks \cite{Liu2001, Rui2004} represent these rules by building an FSM. However, these selection mechanisms have limited scalability to incorporate new semantic cues or rules, and the resultant videos will be mechanical and predictable. To alleviate this problem, Some computational frameworks \cite{Riedl2008, Arev2014, Wu2015, Galvane2015, Leake2017} formulate the editing as an optimization problem and solve it with dynamic programming approaches. Nevertheless, the rigid constraints and the pose-process stage they adopted, such as the shot duration constraint, may result in sub-optimal solutions, and they only perform offline editing. In contrast, this work presents an optimization-based multi-purpose framework with soft constraints to bridge this gap, ensuring optimal solutions and allowing users to adjust video styles and choose different working modes easily.

\section{The proposed system}

\begin{figure}[htbp]
  \includegraphics[scale=0.7]{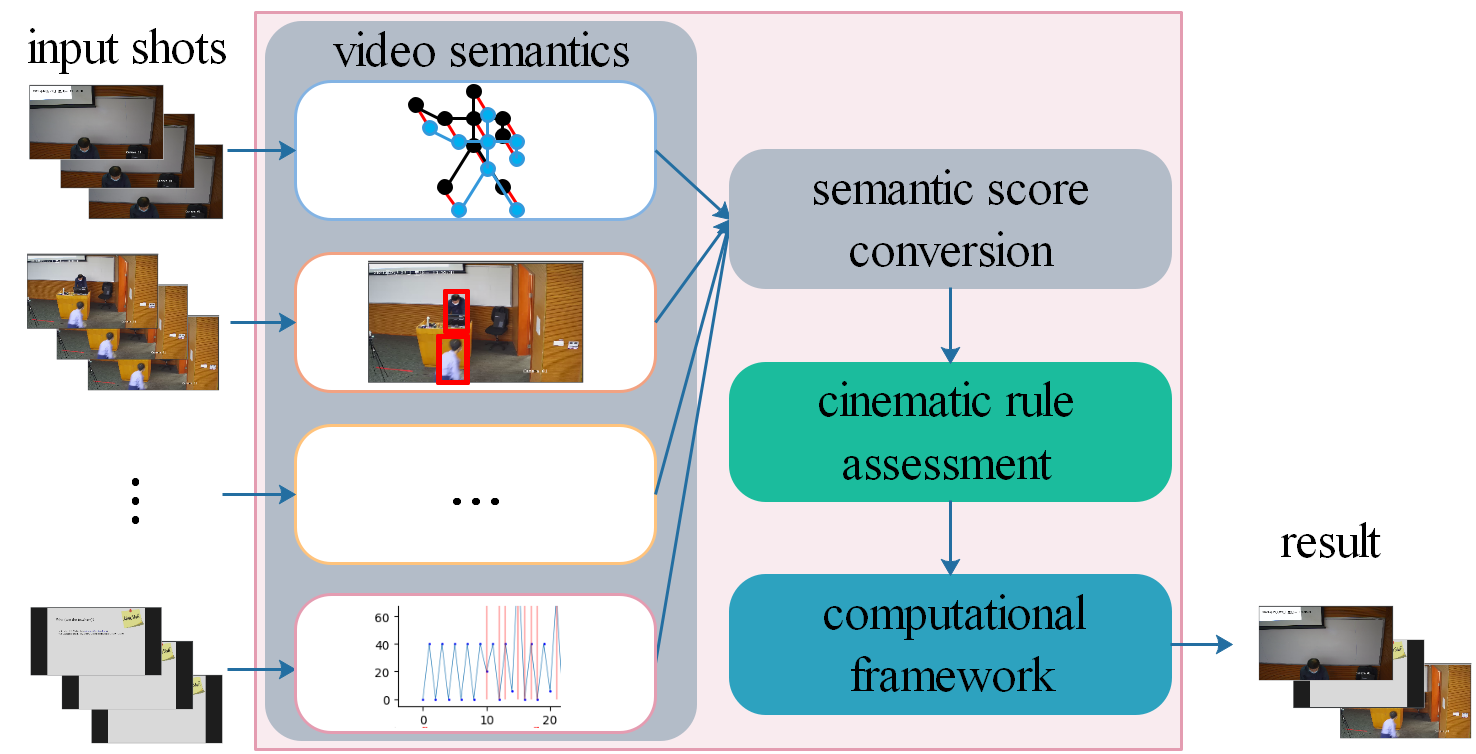}
  \caption{The overall architecture of the proposed editing system.}
  \label{fig:system}
\end{figure}

As reviewed in the previous section, the common shortages in existing lecture broadcast systems mainly include the limitations in understanding the high-level semantics of videos and the weak extendability of the rule-based directing schemes. To tackle these problems, we first propose different semantics extraction methods to assess different shots, which will be discussed in Sec.\ref{sub:incentives}. In Sec.\ref{sub:selection method}, we will introduce our computational editing framework built upon the semantic cues. The overall architecture of our system is illustrated in Fig.\ref{fig:system}, different shots are firstly fed into the independent shot semantics assessment module to generate the event indicators which are then converted to the semantic scores. Next, the semantics scores and the scores from the assessment of general cinematic rules are passed to the computational framework to produce resultant videos.  

\subsection{Problem formulation}
Technically, live broadcasting or editing lecture videos can be regarded as a consecutive view selection process. The inputs to the system include a set of $C$ synchronized video streams $V =\{V_c\}_{c=1:C}$ and each $V_c$ is decoded as a frame sequence $\{f_{c,t}\}_{t=1:T}$ or a clip sequence, it depends on the unit of a time instance. For simplicity, we will use a frame as the unit of time in the rest of this paper. After acquiring $l$ frames starting from time $t$, the system will analyze the content of $\{f_{c,t:t+l}\}_{c=1:C}$ and then select the best views indexed by $\{c_t, c_{t+1},\cdots,c_{t+l}\}$. As a result, the frame sequence $\{f_{c_t,t},f_{c_{t+1}, t+1}, \cdots, f_{c_{t+l},t+l}\}$ are concatenated to form the video stream. For clarification, we may also use the abbreviations of shot names to denote the camera indices or frame sources in this paper. i.e., subscript $lb$ stands for left blackboard close-up shot, $sc$ stands for slide close-up shot, and $sl$ denotes student long shot, etc. It is worth noting that if start time $t=0$ and $l$ is the duration of the lecture, the system will perform the offline editing. On the other hand, if $l$ is set to 0, the system can live broadcast the selected view. In the proposed system, the users can even make a trade-off between these two modes by simply adjusting the duration $l$ looking ahead.




\begin{table}[h]
\centering
\caption{The shot names and the corresponding semantics clues used to assess the focus score. The notions in the brackets are the subscripts indicating the corresponding shots.}
\begin{tabular}{|c|c|}
\hline
 Shot name &  Semantics  \\
\hline
left black close-up shot ($lb$)  &  \multirow{2}{*}{writing event recognition}  \\ 
\cline{1-1}
right black close-up shot ($rb$) & \\
\hline
slide close-up shot ($sc$)&  gradient based anomaly detection  \\ 
\hline
student long shot ($sl$) &  motion entropy difference \\
\hline
left medium shot ($lm$) &  \multirow{2}{*}{the number of detected persons} \\
\cline{1-1}
right medium shot ($rm$)& \\
\hline
overview long shot ($ol$) & the position of speaker \\
\hline
\end{tabular}
\label{tbl:shot semantics map}
\end{table}

\subsection{Shot assessment from video semantics}
\label{sub:incentives}


The first problem to be addressed for video editing is deciding what to show at any given moment \cite{Wang2007}. Generally, a view gaining more attention is assigned with a high score for selection. As shown in Fig.\ref{fig:ourview}, there are seven shots in our systems and the perspectives of these shots are diverse, serving different purposes \cite{Liu2001}. Therefore, our computational editing system assesses the focus of different shots from different aspects. Specifically, we first identify whether a particular event defined for each shot happens at each time point by analyzing its content. The results stored in the indicator vectors are then fused and converted to the focus scores, considering the priorities of shots. Table.\ref{tbl:shot semantics map} summarizes the shot types and the corresponding content semantics used to assess the focus scores.

\noindent \textbf{Blackboard Close-Up Shot (BCUS).} The BCUS contains left BCUS and right BCUS, which are set to capture the content written on the blackboard, and this shot will draw the student's attention when the writing event happens. Hence, the core to assessing this shot is to recognize the writing event. Previous editing systems \cite{Heck2007, Wang2007} detect the writing event by calculating the frame difference over time, while their methods are vulnerable to illumination variation and the movement of the presenter. Skeleton information has proved to be useful information for recognizing human action \cite{Zhu2019,Xia2021,Ng2021}. As shown in Fig.\ref{fig:writing event}, the skeleton topologies for writing events are discriminative from those for non-writing events. It is feasible to recognize these two cases by analyzing the speaker's skeletons. However, existing skeleton-based approaches mostly take as inputs the 3D joint positions that are not acquired from common RGB cameras. Avola et al. \cite{Avola2019} propose a 2D skeleton-based approach that extracts the features of upper and lower body parts with two-branch neural network architecture. Whereas, the lower body part of the presenter is not always visible in the close-up shot.
\begin{figure*}[htbp]
  \centering
  \includegraphics[scale=0.8]{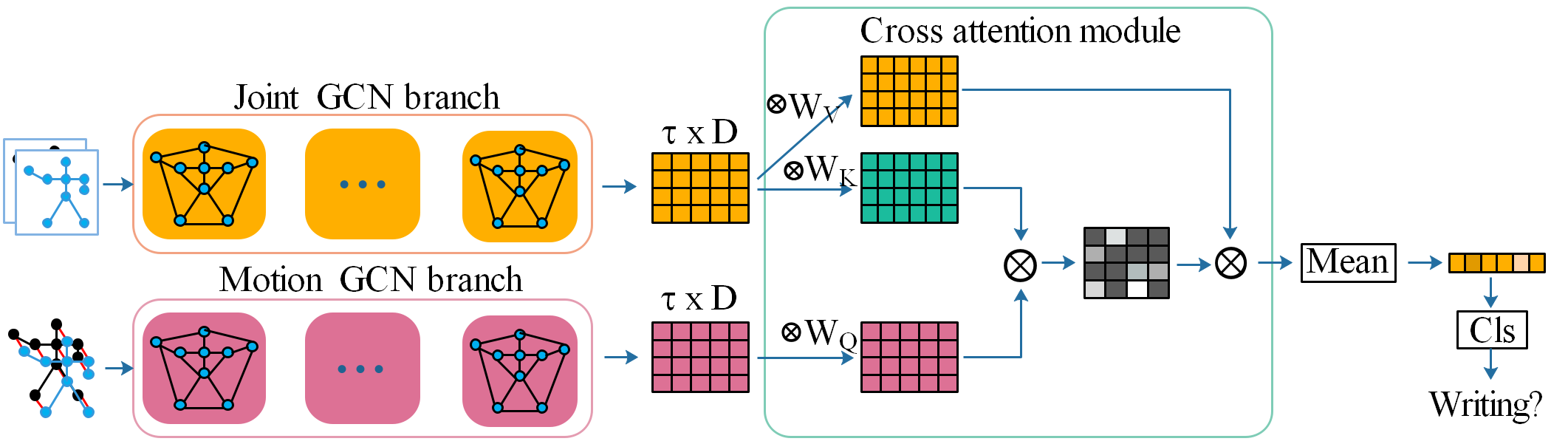}
  \caption{The proposed skeleton-based event recognition architecture, consists of two GCN embedding branches and a cross-attention feature aggregation module.}
  \label{fig:GCNNet}
\end{figure*}

Considering the capability of Graph Convolutional Network (GCN) \cite{Chen2021,Song2022} in representing the topology of the human body, we propose a graph-based cross-attention network to recognize the writing event based on 2D skeletons. As shown in Fig.\ref{fig:GCNNet}, two independent GCN branches extract the static joint feature and the motion features, respectively, followed by a cross-attention block aggregating joint features temporally with attention scores. Specifically, we apply Openpose \cite{Cao2017} to compute 8 joint locations of upper body part $J_{b,t} \in \mathbb{R}^{8\times2}$ for each BCUS frame $f_{b,t}$. To predict whether the writing event occurs at time $t$, a sequence of joint locations $[J_{b,t-\tau},\cdots, J_{b,t}]$, along with the motion sequence $[M_{b,t-\tau},\cdots, M_{b,t}]$ where $M_{b,t} = (J_{b,t}-J_{b,t-\Delta t})/\Delta t$, are fed into joint embedding and motion embedding branches, composed of 5 GCN units \cite{Chen2021} respectively, to compute the joint features $F_{j,t} \in \mathbb{R}^{\tau \times D}$ and motion features $F_{m,t} \in \mathbb{R}^{\tau \times D}$. In the cross attention module, we project $F_{m,t}$ into a query embedding $Q_{t} \in \mathbb{R}^{\tau \times D_p}$ , and $F_{m,t}$ into a key embedding $K_{t} \in \mathbb{R}^{\tau \times D_p}$ and a value embedding $V_{t} \in \mathbb{R}^{\tau \times D_p}$ with three different project matrices $W_{Q}, W_{K}, W_{V} \in \mathbb{R}^{D \times D_p}$:
\begin{align*}
  Q_{t} &= Norm( F_{j,t} )W_{Q} \\
  K_{t} &= Norm( F_{m,t} )W_{K} \\
  V_{t} &= Norm( F_{m,t} )W_{V}
\end{align*}
where $Norm()$ denotes the layer normalization function. Thus, the aggregated feature $F_{ag, t}$ is computed as the average feature vector of the weighted value embedding:
\begin{equation*}
  F_{ag, t} = Mean(Softmax(\frac{Q_{t}K^{T}_{t}}{\sqrt{D_p}})V_{t})
\end{equation*}
Finally, a binary classifier takes $F_{ag, t}$ as input to estimate the probability $p_t$, and an indicator vector for left BCUS, $\mathbf{I}_{lb}$, (right BCUS, $\mathbf{I}_{rb}$) is used to record the event by setting $\mathbf{I}_{lb}[t]$ ($\mathbf{I}_{rb}[t]$) to 1 if $p_{t}$ is greater than a threshold otherwise 0.

\begin{figure}[htbp]
  \centering
  \includegraphics[scale=0.8]{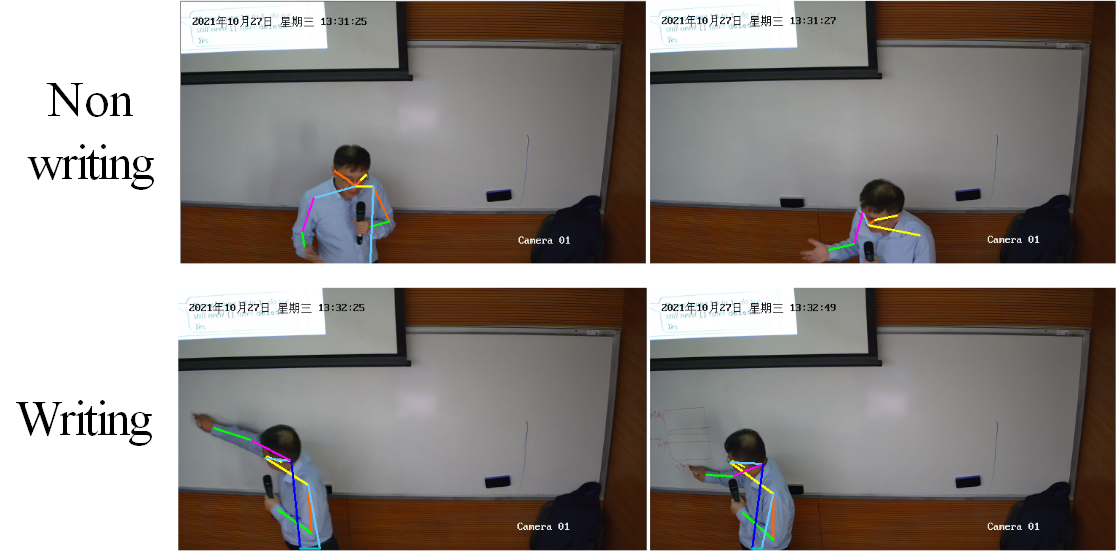}
  \caption{The skeleton topologies for two different situations. The first column is the predicted result by our method. }
  \label{fig:writing event}
\end{figure}

\begin{figure*}[htbp]
  \centering
  \includegraphics[scale=0.7]{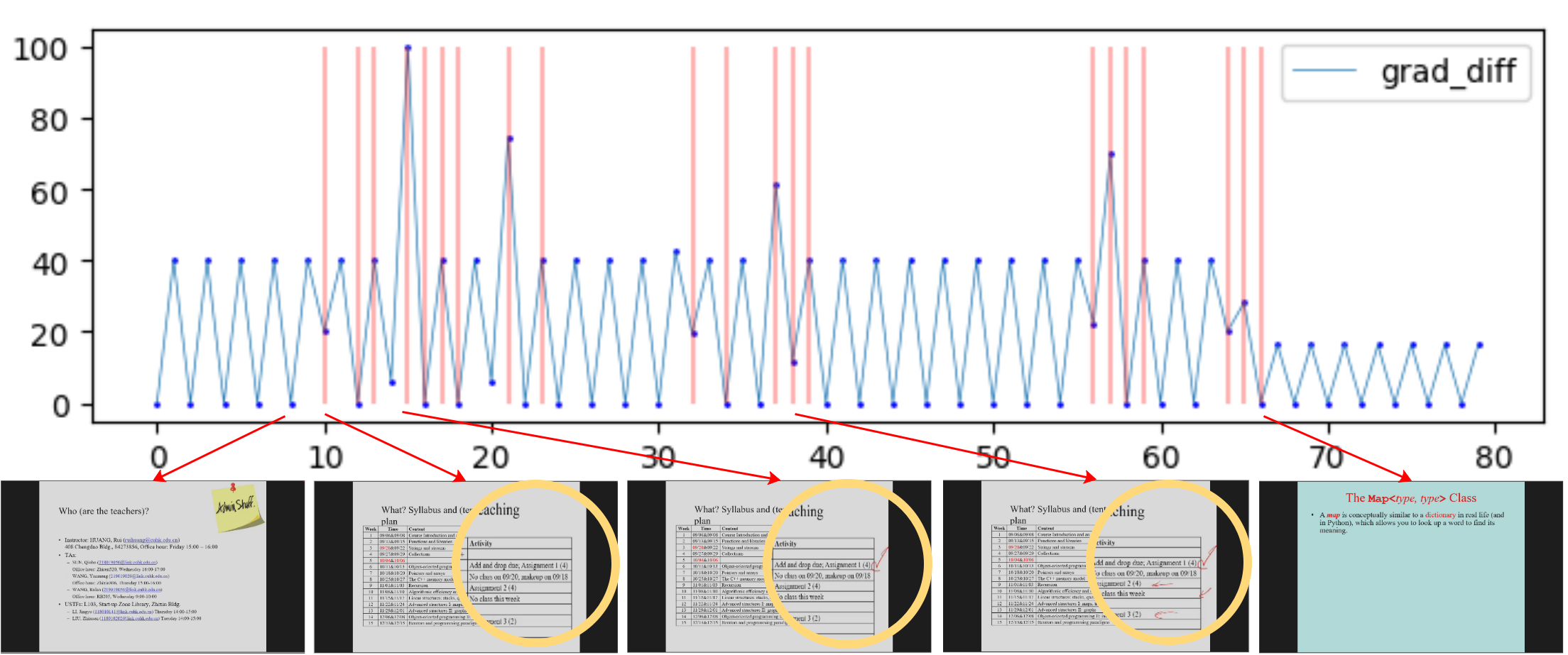}
  \caption{Slide change detection by the proposed anomaly detection method. The blue curve is the gradient difference scores over time, and the red vertical lines are the detected anomalous time points. The bottom row represents the frames sampled from the time points specified by the red arrow line. }
  \label{fig:SlideFlip}
\end{figure*}

\noindent \textbf{Slide Close-Up Shot (SCUS).} Slide projector plays an important role in current classes, teachers use slides to assist their teaching activity. Therefore,  previous editing systems \cite{Wang2007} also take the SCUS into consideration and utilize the gesture and position information of teachers to access its focus from students. Whereas, in a scene such as the large classroom or report hall where the presenter cannot interact with the projected slide but use a laser pointer or the mouse to flip and draw on the slide, the key to assessing the focus is to detect the content changes in the slide. As the color histogram difference method \cite{Rui2004} is susceptible to the video stream noise, and not sensitive to the small streaks drawn by the presenter on the slide, we propose a gradient difference-based anomaly detection method to address the above problems. Let $f_{sc,t-1}$ and $f_{sc,t}$ denote two adjacent frames of slide shot, and the gradient difference score $S_{g,t}$ is calculated as:
\begin{equation*}
  \label{eqn:grad diff}
  S_{g,t} = \frac{1}{3}\sum_{i=1}^{3} \Vert Grad(f_{sc,t-1}[i]) - Grad(f_{sc,t}[i])\Vert_2
\end{equation*}
where $Grad(f_{sc,t-1}[i])$ denotes the function of calculating gradients for  the $i$-th channel of $f_{sc,t-1}$.  To predict whether a salient change occurs on the slide at time $t$,  instead of applying a threshold on score $S_{g,t}$, we employ an autoregressive model-based Anomaly Detector (AD) \cite{Yaacob2010} which is more robust to the stream encoding noises. The AD applies a regressor to learn the autoregressive pattern from historical scores, i.e., $\{S_{g, t-\tau},\cdots, S_{g,t-1}\}$, and identifies $S_{g,t}$ as anomalous if the residual of regression is anomalously large. To reward the selections of SCUS at the anomalous time points, we record them by setting the corresponding elements of the indicator vector $\mathbf{I}_{sc}$ to 1. Thanks to the ability of AD to learn the pattern from historical data, our method still works even if no encoding noise exists. Fig.\ref{fig:SlideFlip} shows the detected results of a video segment. It can be observed that the proposed method can detect the flips to new pages (flip to the second frame and the fifth frame of the bottom row) and the streak changes (the third frame to the fourth frame.), even though the video signals are noised.

\noindent \textbf{Student Long Shot (SLS).} SLS is also important for improving the interest and engagement of edited videos. Generally, in the manual directing scenario, the human director will show the student view when students ask questions in class. Existing systems \cite{Liu2001, Rui2004} use a sound source localization-based technique to locate the talking students, while it requires elaborate device calibration, and its performance is greatly affected by the surrounding noises. From the angle of vision, the salient motion occurring in SLS usually is accompanied by something unusual happening, i.e., a student stands up to raise a question, the student group engages in-class activity, etc. To include these student events, we propose a motion entropy-based anomaly detector to find out the unusual time points of SLS. Specifically, given two adjacent frames of SLS, $f_{sl,t-1}$ and $f_{sl,t}$, we firstly compute the optical flow $[u_{sl,t}, v_{sl,t}]$ between them using the off-the-shelf lite flownet\cite{Hui2018}, and the motion boundary information is encoded as the Histogram Of Gradient (HOG) descriptor $[\mathbf{h}_{u,t}, \mathbf{h}_{v,t}]$. The motion score $S_{m,t}$ is defined as the entropy of HOG feature vectors normalized by the softmax function:
\begin{align*}
  & \tilde{\mathbf{h}}_{u,t}[i] = \frac{e^{\mathbf{h}_{u,t}[i]}}{\sum_k e^{\mathbf{h}_{u,t}[k]} }, \quad \tilde{\mathbf{h}}_{v,t}[i] = \frac{e^{\mathbf{h}_{v,t}[i]}}{\sum_k e^{\mathbf{h}_{v,t}[k]} } \\
  & S_{m,t} = -\sum_{i} \tilde{\mathbf{h}}_{u,t}[i]*\log \tilde{\mathbf{h}}_{u,t}[i] -\sum_{i} \tilde{\mathbf{h}}_{v,t}[i]*\log \tilde{\mathbf{h}}_{v,t}[i]
\end{align*}

\begin{figure}[htbp]
  \centering
  \includegraphics[scale=0.8]{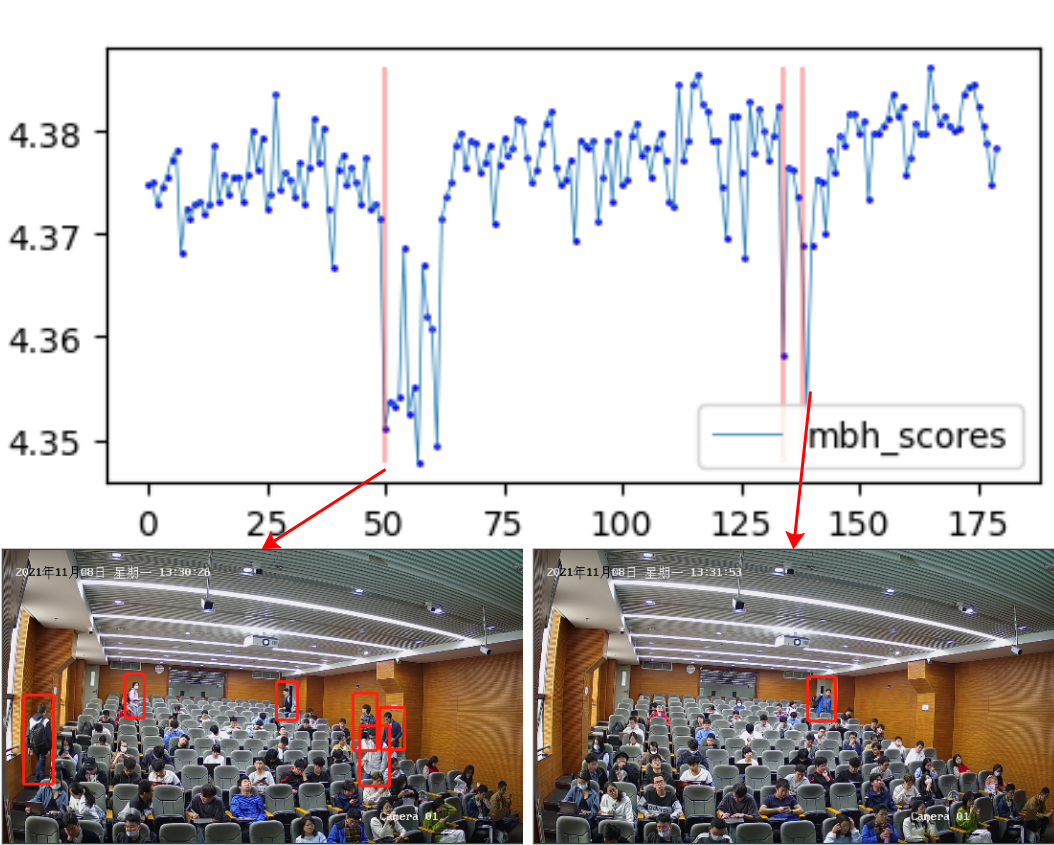}
  \caption{The anomaly detection results for SLS, the blue curve is the sequence of motion scores over time, the red vertical lines denote the detected anomalous time points, and the moving students are highlighted with red bounding boxes.}
  \label{fig:studentevent}
  \vspace{-12pt}
\end{figure}

The motion score curve will drop dramatically as the salient objects move in the same direction. For example, when a student stands up, the pixels of this student will shift up and the left pixels may move slightly in all directions, resulting in a drop in motion score. Finally, we compare the mean score difference between two score windows ,$\{S_{m,t-2w}, c\dots, S_{m, t-w-1}\}$ and $\{S_{m,t-w}, c\dots, S_{m, t}\}$,  with a threshold to identify the anomalous drop, which is considered as an unusual event happening. An indicator vector $\mathbf{I}_{sl}$ is built to record the events by filling the elements at anomalous time points with 1 and the other elements are 0s. Fig.\ref{fig:studentevent} illustrates the detected results for a segment. The motion score curve drops significantly when the salient motions occur, e.g., the students are moving or standing up suddenly.  

\begin{figure}[htbp]
  \centering
  \includegraphics[scale=0.8]{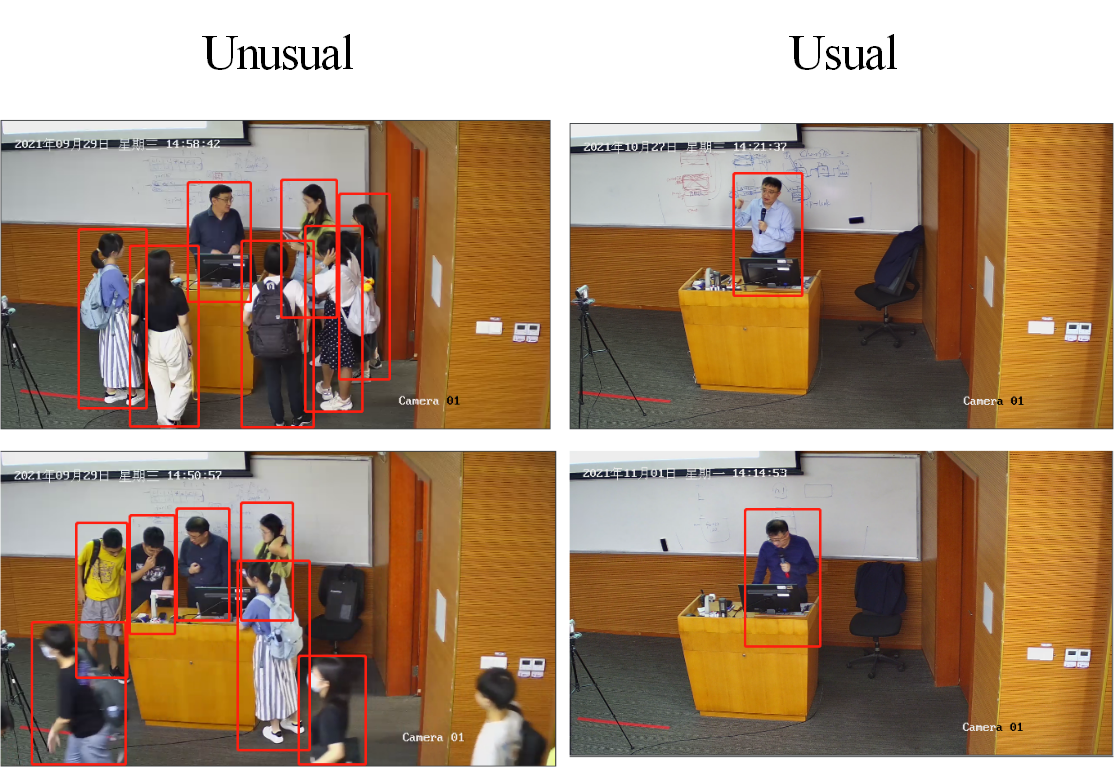}
  \caption{The illustrations of unusual and usual cases for teacher medium shots. The cases where more than one person is detected are identified as unusual, otherwise usual.}
  \label{fig:teacherview}
\end{figure}

\noindent \textbf{Medium Shot (MS).} Our system sets up both the Left Medium Shot (LMS) and Right Medium Shot (RMS) to increase diversity, and they serve the same purpose. The MS is set to capture the whole body of the speaker so that the remote students can keep up with the teacher by watching his gesticulation and the interaction with other students. Similar to other shots, we also define a metric to identify the unusual event. the MS typically contains only the presenter, so it is treated as unusual when more than one person is detected, as shown in Fig.\ref{fig:teacherview}. Hence, we utilize an off-the-shelf detector YOLOv3 \cite{Farhadi2018} to count the number of persons. Let $n_{lm,t}$ ($n_{rm,t}$) denotes the number of detected persons in frame $f_{lm,t} $($f_{rm,t}$), the $t$-th element of indicator vector $\mathbf{I}_{lm}$ ($\mathbf{I}_{rm}$) is set to 1 if $n_{lm,t} > 1$ ($n_{rm,t} > 1$) else $0$.

\begin{figure}[htbp]
  \centering
  \includegraphics[scale=0.8]{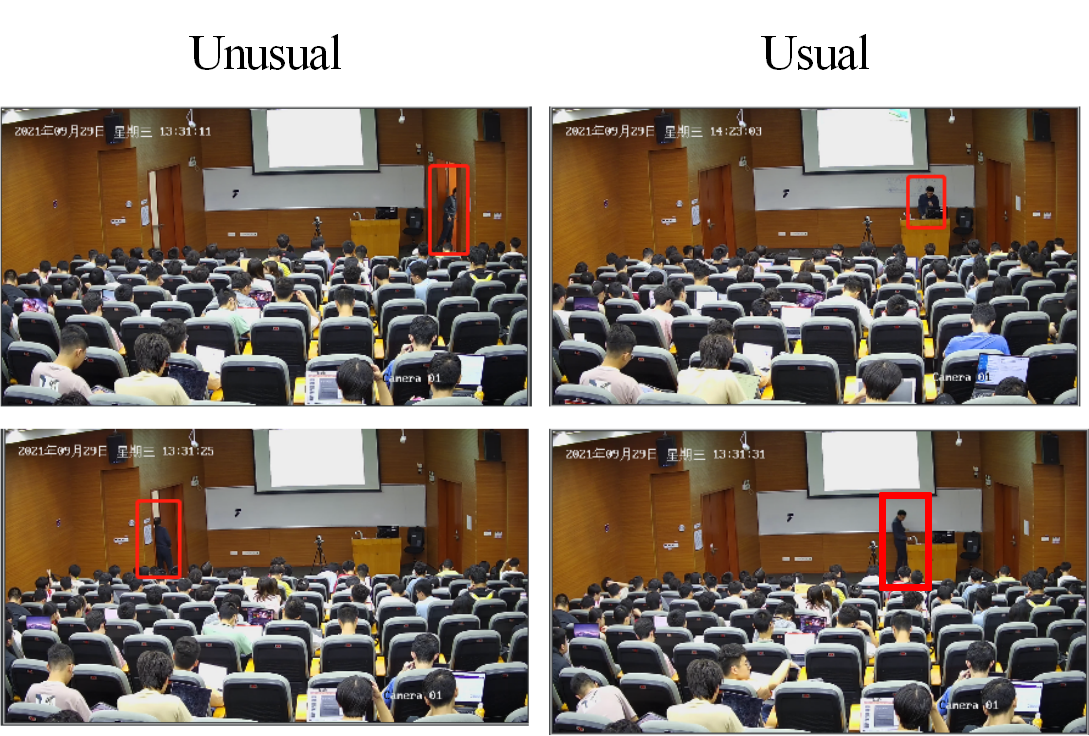}
  \caption{The illustrations of unusual and usual cases for OLS.}
  \label{fig:longshot}
\end{figure}

\noindent \textbf{Overview Long Shot (OLS).} As a complementary shot to the other shots, the OLS can capture the whole classroom and show the presenter's actions happening outside of other shots for the remote students, increasing the engagement and interest of students. As shown in Fig.\ref{fig:longshot}, the speaker usually moves around the podium, and it is considered an unusual case if the speaker exceeds the normal range. To this end, we access this shot by tracking the positions $\{\mathbf{p}_{ol,t}\}_{t=1:T}$ of the presenter over time, and the elements of indicator vector $\mathbf{I}_{ol}$ are set to 1s if the positions at the corresponding time points are greater than a predefined threshold otherwise 0.

\noindent \textbf{Conversion from indicators to scores } Without loss of generality, suppose the current time is $t$ and $l$ frames after $t$ are acquired, we can compute the indicator vectors  $ \mathbf{I}_{rb}, \mathbf{I}_{lb}, \mathbf{I}_{sc}, \mathbf{I}_{sl}, \mathbf{I}_{lm}, \mathbf{I}_{rm}, \mathbf{I}_{ol} $ through the semantics analysis methods mentioned above. As multiple shots may be of interest at the same time, or no shot is of interest at some time points, it needs to weight different shots with the weight vector $[w_{rb}, w_{lb}, w_{sc}, w_{sl}, w_{lm}, w_{rm}, w_{ol}]$ and set up the default score vector $[s_{rb}, s_{lb}, s_{sc}, s_{sl}, s_{lm}, s_{rm}, s_{ol}]$. Therefore, the focus scores for shot selections can be computed with the indicators, weight vector, and default score vector. Take the SCUS as an example, let the camera index of SCUS be $c_i$, and the focus score from semantics for selecting SCUS at time $i$ can be written as:
\begin{equation*}
  r^{e}_{c_i, i} = s_{sc} + \mathbf{I}_{sc}[i]*w_{sc}
\end{equation*} 

The weight vector and the default score vector can be adjusted by the users according to their preferences. Generally, the SCUS and BCUS get higher weights and default scores, thus they are selected as a matter of priority when conflicting with other shots of interest. Acquiescently, the weight vector and default score vector are both set to $[0.8,0.8,1,0.4,0.6,0.6,0.2]$.

\subsection{Shot assessment from cinematographic rules}
\label{sub:rule expression}
Besides video semantics, professional cinematographic rules also have a great impact on the viewing experience \cite{Dmytryk2018}. Unlike the previous systems that hard code the rules, we integrate them by converting the shot selection constraints and suitability into soft computational formulations, which are then used together with the focus scores from video semantics by the proposed optimization-based editing framework to compose the videos. In this way, users can easily adjust the preferred video styles.  

\textbf{View transition constraint.} In professional film editing, there are many empirical constraints \cite{Mascelli1965,Murch2001,Germeys2007} on shot transitions in order to prevent confusing audiences. One fundamental guideline is to avoid \emph{Jump cuts}. This guideline claims that the transition between two camera views that shoot the same scene from almost the same angles e.g., the angle difference is below 30 degrees, will be perceived as a sudden change, resulting in a jarring cut. Unlike the footage from the traditional filming scene or animation scene where the camera angle can change casually, the lecture videos are captured with fixed camera views. Hence, the \emph{Jump cuts} constraint is satisfied in the camera setup stage.

Another core guideline is the \emph{180-degree rule}, which stresses that the cameras of two consecutive shots shooting the same object must situate on one side of an imaginary line-of-action. Otherwise, it will create an abrupt reversal of the action or characters. Similarly,  a rule about the \emph{order of shot} \cite{Aerts2016,Ronfard2021} in shot transitions argues that the shot size should change smoothly, and a common order of shot is to start with a long shot, establishing an overview of the scene. So the shot after a long shot is typically a medium shot, which is then followed by a close-up shot. Although these rules are often pleasing, they are not necessarily always followed. There should be some variation in the sequence to prevent producing too mechanical montages. Therefore, we implement these rules in a soft manner. Specifically, as there are 7 types of shots in our system, we build a $7 \times 7$ matrix $T$, representing the transition suitability of all shot size combinations. The element at position $(c_{start}, c_{end})$ is set as
\begin{equation}
\label{eqn:transition}
  T[c_{start},c_{end}] = \begin{cases}
                           -\epsilon, & \mbox{if } c_{end} \in C_{viol}(c_{start}) \\
                           \epsilon, & \mbox{otherwise}.
                         \end{cases}
\end{equation}
where $C_{viol}(c_{start})$ denotes the set of camera to which the cuttings from $c_{start}$ violate the rules above. For example, if $c_{start}$ is the left close-up shot $c_{lcu}$ and $c_{end}$ is the right close-up shot $c_{rcu}$ or student long shot $s_{ol}$, the transition will violate the \emph{180-degree rule} or the \emph{order of shot} rule. Although the negative element of $T$ is treated as a penalty when multiplied by the semantic score $r^{e}_{c_{end}, t}$, it still leaves the possibility of making such a transition when there are enough incentives from other sources. It is favorable for producing diverse montages.

\noindent \textbf{Switch penalty} Previous works suggested that frequent switches will cause an unpleasant viewing experience while lasting the same view for a long time will make the broadcast tedious. Hence, we dynamically assign a penalty $r^{sw} (\leq 0)$ for each selection based on the duration $L$ that the current view has lasted for.
\begin{equation*}
  r^{sw}(L, switch) = \begin{cases}
                C_{sw}*(\frac{1}{1+e^{(L-L_{max})}}-1), \mbox{if } \sim switch  \\
                C_{sw}*(\frac{1}{1+e^{(L_{min}-L)}}-1), \mbox{if } switch \\
                0,  \mbox{otherwise}.
              \end{cases}
\end{equation*}
where $L_{max}$ and $L_{min}$ represent the expected maximum and minimum segment lengths, respectively. In previous systems \cite{Liu2001,Zhang2008,Arev2014,Wu2015,Aerts2016}, the rigid rules usually force the system to cut the overlong segment or maintain the short segment even if it is meaningless. Instead, our soft penalty mechanism allows the system to remain the same view even if its length has exceeded $L_{max}$ if there are strong incentives from other aspects. For example, the system is not supposed to switch the view where the speaker does not stop writing yet, even though the view has lasted for a long time.

\noindent \textbf{B-roll insertion} Some events in the lecture may last for a long time, e.g., discussion with students. Watching the same view all the time is boring and it may hurt the focus of students. An excellent practice \cite{Liu2001} is to show the B-roll view occasionally for a period of time (e.g., 9 seconds). It will make the resultant video more interesting to watch. A B-roll can be a shot that shows the overview of the classroom $c_{ol}$, the states of student $c_{sl}$, or the teaching materials $c_{scu}$. So we set up an incentive $r^{broll}(\geq 0)$ for inserting B-roll views when current view $c$ has last for a period of time at $t$.
\begin{equation*}
  r^{broll}(L, c_{end}) = \begin{cases}
                           C_{broll},\mbox{if } L > \frac{L_{mean}}{2} \& c_{end} \in \{c_{sl},c_{ol},c_{scu}\} \\
                            0, \mbox{otherwise}.
                          \end{cases}
\end{equation*}
It should be noted that this is not a rigid rule, all the decisions are made during the optimization process. In other words, the B-roll view does not always appear in the optimal solution, even though the triggered conditions are satisfied.

\subsection{Computational editing framework}
\label{sub:selection method}
\begin{figure*}[htbp]
  \centering
  \includegraphics[scale=0.8]{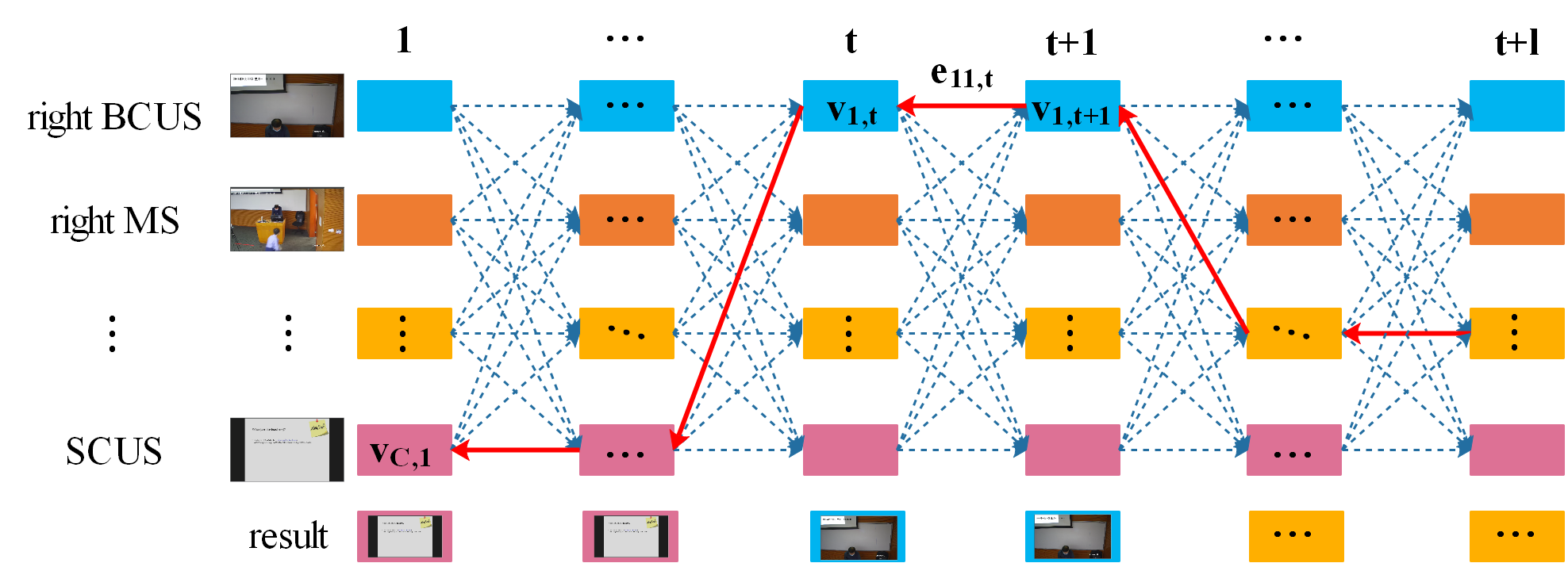}
  \vspace{-12pt}
  \caption{The graph model of the proposed computable editing framework. The first two columns are the shot names and the corresponding views, and the bottom row indicates the resultant video retrieved with the red arrow path.}
  \label{fig:framework}
  \vspace{-12pt}
\end{figure*}

The view selection process is an essential part of the systems. Traditional rule-based frameworks have limited capacities in incorporating new information measurements and cannot balance the real-time performance and the optimality of solutions. Even some systems \cite{Wu2015} have to post-process to prevent over-long and over-short clips. Hence, we propose a complete optimization-based framework that can achieve the optimal solution and enables users to switch the modes, e.g., live broadcasting, offline editing, or a balance between them, by simply adjusting the duration $l$ looking ahead. For example, users can experience live broadcasting by setting $l$ to 0 or obtain the optimal edited video by setting $l$ to the length of the lecture videos or make a trade-off between them. Furthermore, new video semantic cues can be readily embedded by quantizing their importance to each view without re-defining a bundle of rules.

Without loss of generality, we suppose that the selection starts from time $t$ with view $c_t$ which has been lasting for $L_t$ time instances. The information in future $l$ time instances is available as well. The goal of our system is to figure out an optimal view index sequence $s^{*}=\{c^*_{t+1}, \dots, c^*_{t+l}\} \in \mathbf{M}_t$ by solving the following optimization problem, where $\mathbf{M}_t$ is the space of all possible view index sequences from $t+1$ to $t+l$:
\begin{align*}
  \argmax_{\{c_{t+1},\dots,c_{t+l}\}} & \sum_{i=t+1}^{t+l} \lambda_{e} T[c_{i-1},c_{i}] \ast r^{e}_{c_i,i} + \lambda_{b} \ast r^{broll}(L_{i-1},c_{i}) \\
  & + \lambda_{sw} \ast r^{sw}(L_{i-1}, switch) \\
  L_i, switch =& \begin{cases}
           L_{i-1} + 1, False & \mbox{if } c_{i}= c_{i-1} \\
           1, True & \mbox{otherwise}.
         \end{cases}
\end{align*}
where $\lambda_{e},\lambda_{b},\lambda_{sw}$ are the adjustable weights for three reward terms.

Directly applying the brute-force algorithm to search for the optimal solution will result in an exponential complexity $C^{l}$. Instead, we formulate the above optimization problem as a path-searching problem in a directed graph model. The result is solved under the complexity of $l*C^2$. We treat each frame $f_{c,t}$ as a node $v_{c,t}$ in the directed graph, and each edge only exists between two nodes that are temporally adjacent and is directed to the node owning bigger time stamp, e.g., edge $e_{c_1c_2,t+1}$ denotes the edge pointing from $f_{c_1,t+1}$ to $f_{c_2, t+2}$ as shown in Fig.6.

In this graph model, we employ the scheme of the breadth-first search to forward the reward gained by the nodes at time $t$ to those at $t+l$, then backtrack from the node with the maximum rewards to obtain the optimal path. Each node $v_{c, t+i}$ contains three components: the reward gained $R_{c,t+i}$, its precursor $P_{c,t+i}$, and the view length $L_{c,t+i}$. During the forward process, the camera index of $P_{c,t+i}$, the precursor of $v_{c, t+i}$, is found as:
\begin{align*}
  & k* = \argmax_{k \in \{1,\dots,C\}} R_{k, t+i-1} + D_{k,c,t+i}, \quad \text{where} \\
  D_{k,c,t+i} &=\begin{cases}
                      &\lambda_{e}T[k,c]\ast r^{e}_{c,t+i} + \lambda_{b} \ast r^{broll}(L_{k,t+i-1},c) \\
                      &+ \lambda_{sw} \ast r^{sw}(L_{k,t+i-1},False), \mbox{if } k=c \\
                      &\lambda_{e} T[k,c]\ast r^{e}_{c,t+i} + \lambda_{b} \ast r^{broll}(L_{k,t+i-1},c) \\
                      & + \lambda_{sw} \ast r^{sw}(L_{k,t+i-1},True), \mbox{otherwise}.
                    \end{cases}
\end{align*}
Therefore, the component of $v_{c, t+i}$ are updated as:
\begin{align*}
  R_{c,t+i} &= R_{k^*, t+i-1} + D_{k^{*}c,t+i}, \quad
  P_{c,t+i} = v_{k^*,t+i-1}, \quad \\
  L_{c,t+i} &= \begin{cases}
                 L_{k^*,t+i-1} + 1, & \mbox{if } k^*=c \\
                 1, & \mbox{otherwise}.
               \end{cases}
\end{align*}
As all the rewards  $\{R_{k,t+l}|k=1,\dots,C\}$ of the nodes $\{v_{k,t+l}|k=1,\dots,C\}$ at time $t+l$ have been obtained, we will trace the path backward from time $t+l$. The camera index sequence $\{c^*_{t+l}, c^{*}_{t+l-1},\dots, c^{*}_{t+1}\}$ is:
\begin{align*}
  c^{*}_{t+l} &= \argmax_{k \in \{1,\dots,C\}} R_{k,t+l}, \quad
  c^{*}_{t+l-1} = Cam(P_{c^{*}_{t+l},t+l}), \quad \\
   & \cdots, \quad
  c^{*}_{t+1} = Cam(P_{c^{*}_{t+2},t+2})\\
\end{align*}

In the forward process of this framework, each node is updated up to $C$ times at most, and there are $C*l$ nodes, so the solution is derived under the complexity of $C^2*l$.

\section{Experiments}
Although video editing systems have been widely studied, evaluating such systems is still an open problem. The difficulty of assessing video editing systems can be traced to at least three reasons \cite{Lino2014}: 1) There is never a single correct answer to editing problems. Even if the annotators are sophisticated film experts, the solutions they produce may be very different from each other. 2) The editing quality cannot be measured directly since the editing effects are often invisible. 3) The rules of good editing are not absolute. They can guide the editing but are not always strictly followed by the filming experts. Probably because of these reasons, we find no public datasets to measure the progress of this field.


Comparing the predicted solutions with the ground truth might not be feasible yet, but the researchers can still evaluate the editing system from some other aspects, such as optimality,  extendability, ease of implementation, etc., as suggested in the study \cite{Lino2014}. To this end, we collected a set of lecture video data with our recording system from 10 actual classes. There are seven camera views in total, as shown in Fig.\ref{fig:ourview}, and the average length of each view of each class is about 50 minutes, so the total length of the videos is about 3500 minutes. To train the proposed writing event recognition network, We manually annotate the time points when the writing event occurs and use one-quarter of the data for training while the rest is used for testing. In the following section, we propose a set of metrics used to quantitatively measure the properties of videos, thus different algorithms can be compared by inspecting the properties of the generated videos in Sec.\ref{subsec: comparison}. Besides, we conduct a user study to collect and analyze the real user experience in Sec.\ref{subsec:user study}.

\subsection{Comparisons}
\label{subsec: comparison}
Firstly, we compare the outcomes of our system \textbf{Optim($l$)}, where $l$ is the duration look-ahead, with those from the other four methods under our experimental environment: 
\begin{enumerate}
    \item \textbf{Randseg($n$)}\cite{Otani2019}, which randomly selects the segment with length $n$;
    \item \textbf{Ranking} \cite{Saini2012}, which greedily selects the view with the highest event rewards when the current shot length reaches the sampled length from a normal distribution;
    \item \textbf{FSM} \cite{Liu2001, Rui2004}, where the states and the transitions are defined based on our environment;
    \item \textbf{Cons-Optim}\cite{Arev2014}, which is a constrained optimization-based method. 
\end{enumerate}
We set the expected maximum and minimum shot length, $L_{max}$ and $L_{min}$, to 60 and 20, respectively, the mean length and variance for \textbf{Ranking} to $(L_{max}+L_{min})/2$ and 10 seconds, and the rewards weights $\{\lambda_{sw}, \lambda_e, \lambda_b\}$ are set to $\{0.4,0.3,0.3\}$. Actually, the proposed framework allows the users to set up the parameters according to their preferences to generate productions with varied styles. The impacts of these parameters will be studied in Sec.\ref{subsec:abl study}.

Table.\ref{tbl:comparisons} lists the results of four metrics that reflect the properties of editing productions. The experiments with \textbf{+GT} are conducted with the ground-truth writing event annotations, while the left experiments use the predicted event results. The metrics used are defined as follows:
\begin{enumerate}
    \item $R_{avg}$: the average focus score gained by taking the generated shot index sequence, using all of the proposed focus score terms;
    \item $r_{trans}$: the ratio of favorable transitions as discussed in Sec.\ref{sub:rule expression}. Suppose $c_{start}, c_{end}$ are the indices of two different consecutive shots (the transition from $c_{start}$ to $c_{end}$), it is a favorable transition if $T[c_{start}, c_{end}]>0$. $r_{trans}$ measures the ratio of favorable transitions over all transitions;
    \item $r_{max}$: the percentage of frames with the highest focus score of semantics at their own time points; Suppose the camera index sequence of the generated video is $[c_0,c_1, \cdots, c_T]$, $r_{max}$ is computed as:
    \begin{equation*}
        r_{max} = \frac{\sum_t \mathbb{I}(\argmax_{c} r^{e}_{c,t} = c_t)}{T}
    \end{equation*}
    where the function $\mathbb{I}(\cdot)$ returns 1 if the condition in the brackets is satisfied otherwise 0. This metric measures the importance of semantics focus score in video generation to some extent;
    \item $n_{sw}$: the average number of cuts;
    \item $L_{avg}$: the average shot length, where a shot is a sequence of consecutive frames from the same camera.
\end{enumerate}    

\begin{table}[h]
\centering
\caption{The statistics of the editing productions from different methods, and the best results are highlighted in \textbf{bold type}.}
\begin{tabular}{l|c|c|c|c|c}
\hline
 Method & $R_{avg}$ & $r_{max}$& $r_{trans}$& $n_{sw}$ &$L_{avg}$  \\ \hline
\textbf{Randseg(30)}  & 25.2  & 11.4 \%  &56.2 \% & 104 & 33.5  \\
\textbf{FSM}        & 66.7 & 56.0 \%  &97.2 \% & 143 & 24.4 \\
\textbf{Ranking}    & 51.9 & 40.0 \%  &71.6 \% & 87 & 39.9 \\
\textbf{Cons-Optim}  & 68.2 & 52.6 \% & 100 \% & 95 & 36.6 \\

\textbf{Optim($1$)}       & 65.7  & 53.8 \%  &100.0 \% & 178 & 19.6\\
\textbf{Optim($\frac{L_{min}}{2}$)} & 72.2  & 63.1 \% & 98.7 \% & 153 & 22.8 \\
\textbf{Optim($L_{min}$)}     & 72.4  & 63.3 \%  & 98.6 \% & 140 & 24.9 \\
\textbf{Optim($\infty$)}  & 72.7  & 62.5 \%  & 100 \% & 130 & 26.8 \\
\hline
\textbf{Randseg(30)+GT}  & 28.6 & 9.4 \%  & 59.2 \% & 118 & 29.5 \\
\textbf{FSM+GT}        & 96.8 & 71.3 \%  & 89.1 \% & 166 & 21.0 \\
\textbf{Ranking+GT}    & 66.9 & 43.9 \%  & 71.1 \% & 89 & 39.4  \\
\textbf{Cons-Optim+GT}  & 108.9 & 78.3 \%  & 97.6 \% & 84 & 41.3 \\
\textbf{Optim($1$)+GT}  & 109.5  & 79.2 \%  & 100.0 \% & 138 & 25.2 \\
\textbf{Optim($\frac{L_{min}}{2}$)+GT}    & 116.9 & 88.2 \%  &99.2 \% & 129 & 27.0 \\
\textbf{Optim($L_{min}$)+GT}     & 118.0  & 88.3 \%  &99.2 \% & 121 & 28.8 \\
\textbf{Optim($\infty$)+GT}  & 119.1  & 88.5 \%  & 99.1 \% & 115 & 30.3 \\
\hline
\end{tabular}
\label{tbl:comparisons}
\end{table}

From the listed results, our editing model \textbf{Optim($l$)} attains the highest rewards compared to the other four methods in both setups, except for our online mode \textbf{Optim($1$)} which obtains lower reward 65.7 than \textbf{Cons-Optim} when using predicted events. These results prove that our editing model can achieve optimal solutions. Unlike previous methods which heuristically select the shots, our system formulates all the editing rules or constraints into computational expressions, which are further integrated into a unified framework, so the resultant videos are globally optimal. Furthermore, as the look-ahead duration $l$ increases, the scores gained increase accordingly, because more information can be used during the optimization process. As a result, the offline editing mode \textbf{Optim($\infty$)} achieves the highest reward as expected, and the online $l=1$ editing production obtains lower rewards.

In addition, the maximum rates $r_{max}$ of \textbf{Optim($l$)} are relatively higher than those of other methods. It means that the productions show more views favored by the events at the corresponding time points. Moreover, our methods \textbf{Optim($l$)} (and \textbf{Optim($l$)+GT}) cause less transition errors, comparing with other methods in $r_{trans}$, while the average shot lengths of our methods still satisfy the empirical shot length constraints ($L_{min}$ and $L_{max}$) except for the online mode \textbf{Optim($1$)}. It further certifies that our system can make a good balance between different editing rules and is superior to the traditional rule-based editing/broadcasting methods.

Our system can work in different modes by adjusting the looking-ahead duration $l$, and we also study the impacts of $l$ in our system as shown in Table.\ref{tbl:comparisons}. To avoid the interference of inaccurate event recognition, we compare the results of \textbf{Optim($l$)+GT}, and the results are diverse in terms of the proposed metrics. As $l$ increases, the scores gained $R_{avg}$, maximum rate $r_{max}$, and average shot length $L_{avg}$ show a increasing trend, while the favorable transition rate $r_{trans}$ decrease gradually. The larger $l$ means more future information can be used in the optimization process, so the system tends to find the path that attains a higher semantics score, even if penalized by the transition errors and switch penalty since the semantics focus score is the main source of score gained while other constraints are the penalties. In contrast. For the small $l$, the system will focus more on avoiding the penalties, as it does not know whether violating the constraints gains more semantics scores in the future. 

\subsection{User study}
\label{subsec:user study}
\begin{table}[h]
\centering
\caption{The scores of three methods on six questions.}
\begin{tabular}{|c|c|c|c|c|c|c|}
\hline
 Question & 1 & 2 & 3 & 4 & 5 & 6 \\
\hline
\textbf{Zoom} &  3.0 & 2.95 & 3.6  & 3.4  & \textbf{4.25}   & 3.4\\ 
\textbf{FSM} &  3.45 & 3.1 &  3.05  & 3.0  & 2.75    &2.95\\ 
\textbf{Ours} &  \textbf{4.2}  & \textbf{4.05} &  \textbf{3.9}  & \textbf{4.05}  & 3.35    & \textbf{4.0}\\ 
\hline
\end{tabular}
\label{tbl:user study}
\end{table}

As mentioned before, objectively evaluating videos or editing systems is still an open problem, the construction of the ground-truth videos and the evaluation metrics are still understudied. Therefore, We also assess the system from the aspect of real user experience and conduct a user study to evaluate the qualities of the generated videos. Specifically, we recruited 20 volunteers, including 12 undergraduates and 8 postgraduates, and randomly show them the videos generated by three algorithms:1) \textbf{Zoom}, which greedily selects the BCUS or the  based on the writing event or slide flip, simulating the scenario of the popular online teaching software, Zoom, where only two views are available; 2) \textbf{Ours}, the proposed system; 3) \textbf{FSM}, we follow the works \cite{Liu2001, Rui2004} to implement an FSM based editing system under our experimental scene. After watching the videos, the volunteers are asked to score the videos from 1 to 5 with respect to six questions:
\begin{enumerate}
    \item \emph{Do you feel the experience of taking class onsite when watching this video?}
    \item \emph{Is this video interesting and having a pleasing viewing experience?}
    \item \emph{Do you think the shots are selected appropriately according to the semantics of different shots?}
    \item \emph{Is this video effective and helpful to study the course if you are taking the course for the first time?}
    \item \emph{Is this video effective and helpful to review the course?}
    \item \emph{what overall score you will assign to this video? }
\end{enumerate}

The average scores for these questions are summarized in Table.\ref{tbl:user study}. From the scores of the first two questions, the multi-shot algorithms, \textbf{FSM} and \textbf{Ours} achieve higher scores than the two-shot algorithms \textbf{Zoom}, which prove that occasionally displaying some other perspectives of the classroom besides two conventional shots (BCUS and SCUS) can increase the interest of the video and improve their educational experience in taking the online course. According to the scores for the third question, our system can respond to various class semantics and select the shots more appropriately. Moreover, the scores of the proposed method are higher than those of \textbf{FSM} on all six aspects, and than the scores of \textbf{Zoom} on five questions except for the fifth question, which justifies that the videos generated with the proposed editing framework are more attractive and appreciated by the students. However, when it comes to the review purpose, \textbf{Zoom} obtains a higher average score, since its resultant videos are composed of only two shots, allowing the students to locate the content quickly. This result also suggests the importance of the capability in generating diverse videos according to user's preferences and the flexibility in running on different modes.

\begin{figure*}[htbp]
  \centering
  \includegraphics[scale=0.8]{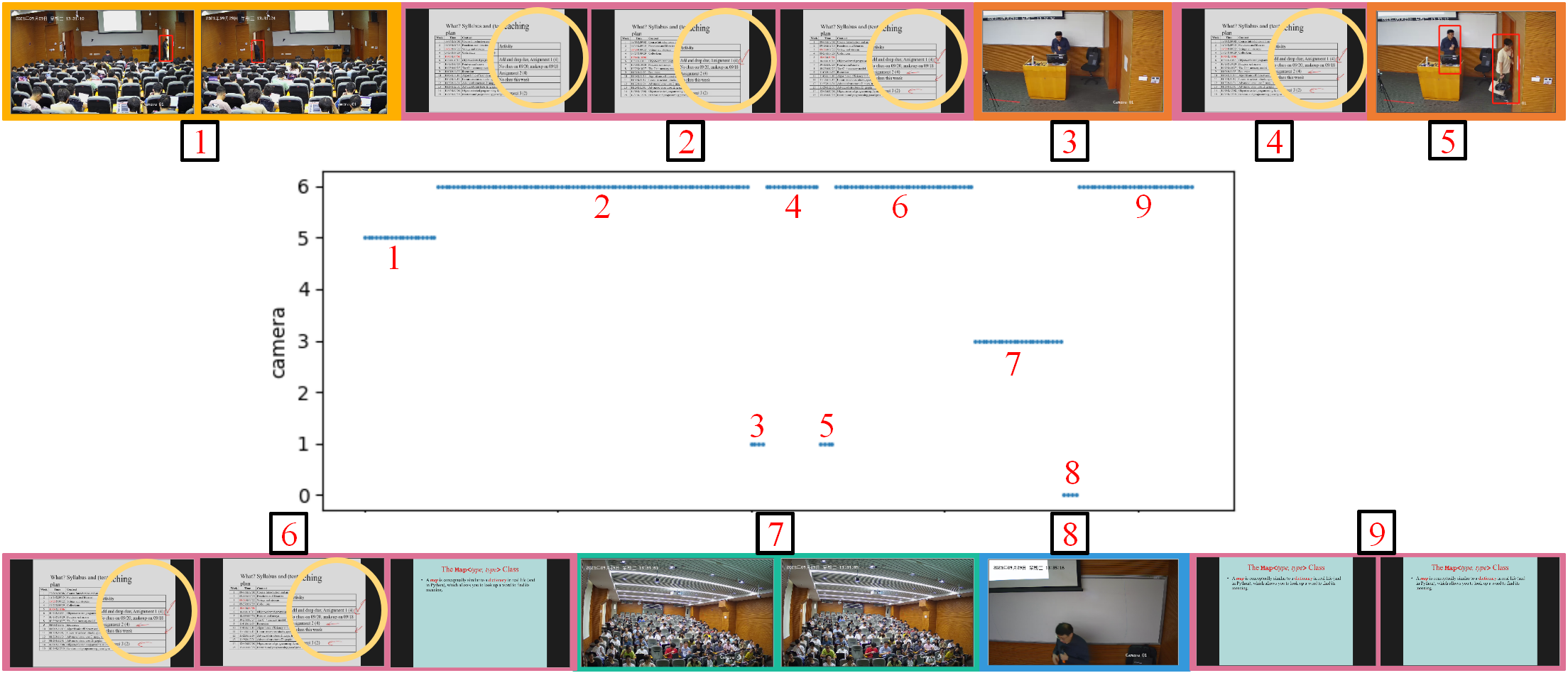}
  \caption{9 consecutive segments of an edited video are visualized. The central figure illustrates the temporal orders and the lengths of these segments. The up row and the bottom row show the sampled frames from these 9 segments associated with the numbers.}
  \label{fig:visualization}
\end{figure*}

\subsection{Visualization results}
To intuitively observe the performance of the proposed system, we visualize 9 consecutive segments of a resultant video in Fig.\ref{fig:visualization}. The temporal relationships and the segment lengths are visualized in the central figure where the vertical and the horizontal axes represent the camera indices and the timeline, respectively. The selection process is inspected as follows:
\begin{enumerate}
    \item Segment 1: The teacher closes the door on two sides, his position is out of the normal region, so the system selects the Overview Long Shot (OLS) to display what happens;
    \item Segment 2: Something changed appears on the slide, so the Slide Close-Up Shot (SCUS) is chosen to show the content details and the annotations from the teacher.
    \item Segment 3: The successive changes on slides result in the system staying on the slide view for a long duration, so the b-roll shot, i.e., the Medium Shot (MS) is selected and lasts for a few seconds. It can prevent the segment of the resultant video from exceeding the maximum shot length while relieving visual tension.
    \item Segment 4: After showing the b-roll shot, the system switches back to the SCUS.
    \item Segment 5: At this moment, multiple people are detected in MS while no other event happens, so the system switches to the MS as expected and lasts for a few seconds.
    \item Segment 6: After a while, the teacher flips the slides, so SCUS is chosen.
    \item Segment 7: The teacher asks the student a question, and rapid motion change is detected in the Student Long Shot (SLS) as the students raise their hands up, so the system switch to the SLS to the responses from the students.
    \item Segment 8: The system should have switched to the SCUS for gaining more scores when there is no special event detected. Whereas directly switching from SLS to SCUS violate the predefined transition constraints, the system instead selects Blackboard Close-Up Shot (BCUS) to avoid the penalty. It is worth noting that the transition constraints vary from scene to scene, this example mainly proves that our editing framework is capable of dealing with varied constraints effectively.
    \item Segment 9: After the BCUS, the SCUS is selected without penalty from the transition constraints.
\end{enumerate}
According to these observations, it can be concluded that our system can edit the videos based on the semantics of lectures while following the general filming rule to ensure a pleasant viewing experience.

\subsection{Ablation study}
\vspace{-12pt}
\label{subsec:abl study}
\subsubsection{The impacts of the score weights}
\begin{table}[h]
\centering
\caption{The experimental results with varied score weights. $+$ and $-$ indicate the experiments with and without transition constraints, respectively.}
\begin{tabular}{c|c|c|c|c|c}
\hline
 trans. constrain&$\{\lambda_{sw},\lambda_e, \lambda_b\}$  & $r_{max}$ &$r_{trans}$& $n_{sw}$ &$L_{avg}$ \\ \hline
\multirow{5}{*}{$+$}  &$\{0.4, 0.3, 0.3\}$ & 88.5 \% & 99.1 \% & 115 & 30.3\\
&$\{1,0,0\}$  & 11.7 \%  & 100 \% & 57 & 60.6 \\
&$\{0.5, 0.5, 0\}$   & 83.3 \%  & 98.1 \% & 106 & 32.8 \\
&$\{0, 1, 0\}$   & 99.9 \%  & 100 \% & 208 & 16.8 \\
&$\{0, 0.5, 0.5\}$   & 99.3 \%  & 100 \% & 254 & 13.7  \\
&$\{0.5, 0, 0.5\}$   & 17.9 \%  & 100 \% &  86 & 40.4\\
\hline
\multirow{5}{*}{$-$}  &$\{0.4, 0.3, 0.3\}$   & 88.8 \%  & 98.3 \% &  116 & 30.3\\
&$\{1,0,0\}$ &11.7 \%  & 100 \% & 57 & 60.6 \\
&$\{0.5, 0.5, 0\}$   & 83.7 \%  & 96.3 \% & 108 & 32.2 \\
&$\{0, 1, 0\}$   & 100 \%  & 98.5 \% & 205 & 17.0 \\
&$\{0, 0.5, 0.5\}$   & 99.3 \%  & 98.8 \% & 251 & 13.9  \\
&$\{0.5, 0, 0.5\}$   & 17.9 \%  & 100 \% &  86 & 40.4 \\
\hline
\end{tabular}
\label{tbl:abl on weights}
\end{table}

In addition to optimality, one advantage of our system is its flexibility, enabling the users to incorporate various information measurements without laboriousness to define the selection rules. Besides, users can adjust the weights of various measurements based on their preferences to generate various productions. In this section, we will discuss the impact of each score term by adjusting its weights. All the experiments are carried out with \textbf{Optim($\infty$)+GT}, and the results have been summarized in Table.\ref{tbl:abl on weights}.

We firstly validate the effectiveness of transition constraint by comparing $r_{trans}$ of the experiments with ($+$) or without ($-$) this term. It is easy to observe that the results with this constraint are usually better than those without it. For example, the experiment with $\{\lambda_{sw},\lambda_e, \lambda_b\} = \{0.4, 0.3, 0.3\}$ achieve higher $r_{trans}$ when applying the transition constraint. With some particular parameters, e.g. $\{\lambda_{sw},\lambda_e, \lambda_b\} = \{0.5, 0, 0.5\}$, the experiments achieve the same $r_{trans}$ on both sides, the reason is that the transition constraint works on the semantic score as a multiplier, so it will be ineffective if semantics score is zero. Moreover, as the weight $\lambda_{sw}$ of switch penalty increases from 0 to 1, $L_{avg}$ also rises up accordingly from 16.8 to 60.6. It concludes that the shot length constraint can be satisfied as the weights incline to switch penalty. In contrast, if a larger weight $\lambda_s$ is put on the semantic score, the maximum rate $r_{max}$ also gets improved, while it will lower the importance of the switch. These observations not only validate the flexibility of our system and its strength in balancing different score items but further show the capacity of our system to generate diverse productions.

\subsubsection{Writing event recognition}
\begin{table}[h]
\centering
\caption{The writing event recognition performances of the traditional method and the proposed methods.}
\begin{tabular}{c|c|c|c|c|c}
\hline
 Method & Accuracy & AUC & Recall & Precision & F1 \\
\hline
\textbf{SVM} &  67.0 & 64.8  & 25.4  & 67.2  & 36.9  \\ 
\textbf{Ours} &  68.8  &  69.3  & 48.1  & 59.6  & 53.3  \\ 
\hline
\end{tabular}
\label{tbl:recognition performance}
\end{table}

\begin{table}[h]
\centering
\caption{The comparisons of editing results with different writing event inputs.}
\begin{tabular}{c|c|c|c|c|c}
\hline
 Method & $R_{avg}$ & $r_{max}$ &$r_{trans}$& $n_{sw}$ &$L_{avg}$ \\
\hline
\textbf{SVM} &  58.3  & 57.0 \% &100 \% & 140 & 24.9 \\ 
\textbf{Ours} &  72.7  &  62.5 \% & 100 \% & 130 & 26.8  \\ 
\hline

\textbf{GT} &  119.1  & 88.5 \% & 99.1 \% & 115 & 30.3\\
\hline
\end{tabular}
\label{tbl:recognition}
\end{table}

As discussed in Sec.\ref{sub:incentives}, we propose a skeleton-based two-stream GCN architecture for discriminating the writing event from the non-writing event. In this section, we will compare it with the traditional SVM method and study the impacts of on the editing system. All the experiments are conducted with \textbf{Optim($\infty$)} where $\{\lambda_{sw},\lambda_e, \lambda_b\} = \{0.4, 0.3, 0.3\}$. Table.\ref{tbl:recognition performance} shows the recognition performances of two methods, and the proposed method outperforms SVM in terms of accuracy, Recall, F1 score, and the Area Under the Curve (AUC). To validate their effectiveness, we apply the predicted results to our editing system, and the results are listed in Table.\ref{tbl:recognition}. By using our recognition method, the editing results achieve higher $R_{avg} = 72.7$, which surpass the $R_{avg}=58.3$ of the results generated with SVM predictions. $r_{max}$ also increases with the help of our method. These comparisons prove the superiority of our method over the traditional method, although there is still a large gap between the predicted results from our method and the ground truth. This experiment also suggests that the class semantics analysis is still under study, and more efforts are needed to promote the development of remote education.


\section{Conclusion and discussion}
\noindent \textbf{Conclusion.} To enhance the educational experience of mixed-mode teaching, we present a multi-purpose semantics-based editing system to live broadcast or offline edit lecture videos for remote students. Beyond the traditional systems using the low-level editing cues and the rule-based selection scheme, we exploit the skeleton of the teacher and formulate the filming rules or constraints into computational expressions, which are integrated into our optimization-based framework to achieve optimal solutions. Both quantitative and qualitative experiments have been conducted to validate the effectiveness of the proposed incentives and the optimality and flexibility of the whole system.

\noindent \textbf{Discussion.} Although our system has made obvious progress in this area from the experimental results and the user study, it still can be improved in a few aspects. We all know that different students may have different watching preferences, which means that the hyper-parameters involved and even the focus measurements are different from person to person. Therefore, the viewing experience can be further improved if the editing system can learn the customized parameters and measurements for each student from his/her own watching behavior, i.e., the customized shot sequence composed on his/her own. Hence, a potential direction is to study learning-based editing techniques, and thus the editing agent can imitate the customized watching behavior and generate the customized videos, after watching a few or even one example, i.e., one-shot imitation learning-based video editing.

\bibliographystyle{IEEETran}
\bibliography{main}

\end{document}